\documentclass[preprint,a4paper,10pt]{elsarticle}
\usepackage{epstopdf}
\usepackage{caption}
\usepackage{setspace}
\usepackage{amssymb}
\usepackage{multirow}
\usepackage{amsthm}
\usepackage{booktabs}
\usepackage{amsmath}
\usepackage{amssymb}
\usepackage{mathrsfs}
\usepackage{longtable}
\usepackage{lscape}
\usepackage{url}
\usepackage{tabularx}
\usepackage{epsfig}
\usepackage{colortbl}
\usepackage{subfigure}
\usepackage{tikz,mathpazo}
\usepackage{graphicx}
\usepackage{natbib}
\usepackage[noend]{algpseudocode}
\usepackage{algorithm}
\usepackage{algorithmicx}   
\usepackage{array}  
\usepackage{hyperref}
\usepackage[figuresright]{rotating}
\graphicspath{{figs/}{figsgaoerb/}}

\newtheorem{myDef}{Definition}

\usetikzlibrary{shapes.geometric, arrows}
\journal{Information Fusion}

\begin{document}
	\begin{frontmatter}
		\title{Construction Cost Index Forecasting: A Multi-feature Fusion Approach}
		\author[address1,address2]{Tianxiang Zhan}
		\author[address1,address2]{Yuanpeng He }
		\author[address2,address3]{Fuyuan Xiao \corref{mycorrespondingauthor}}
		
		\address[address1]{College of Computer and Information Science College of Software, Southwest University, Chongqing, 400715, China}
		\address[address2]{School of Big Data and Software Engineering, Chongqing University, Chongqing, 401331, China}
		\address[address3]{National Engineering Laboratory for Integrated Aero-Space-Ground-Ocean Big Data Application Technology, China}
		\cortext[mycorrespondingauthor]{Corresponding author: Fuyuan Xiao is with the School of Big Data and Software Engineering, Chongqing University, Chongqing 401331, China. (e-mail: xiaofuyuan@cqu.edu.cn; doctorxiaofy@hotmail.com)
		}
	
		\begin{abstract}
		The construction cost index is an important indicator of the construction industry. Predicting CCI has important practical significance. This paper combines information fusion with machine learning, and proposes a multi-feature fusion (MFF) module for time series forecasting. Compared with the convolution module, the MFF module is a module that extracts certain features. Experiments have proved that the combination of MFF module and multi-layer perceptron has a relatively good prediction effect. The MFF neural network model has high prediction accuracy and efficient prediction efficiency. At the same time, MFF continues to improve the potential of prediction accuracy, which is a study of continuous attention.
		\end{abstract}
	
	
	\end{frontmatter}
	
	\section{Introduction}
	The construction cost index (CCI) is an indicator that reflects the construction cost, and it is a research hotspot in the fields of construction and finance. The prediction of CCI is meaningful and necessary. Effectively improving the prediction level of CCI is one of the research goals. CCI data is a time series, and there are many forecasting methods for time series. Time series forecasting methods include statistical methods, fuzzy forecasting methods \cite{parida2017times,Xiao2020maximum,xiao2021caftr,xie2021informationquality}, complex methods \cite{CHEN2021104438,Xiao2021CEQD}, evidence theory methods \cite{https://doi.org/10.1002/int.22598}, machine learning methods \cite{zhou2021informer,huang2021new,HUANG2021107478}, deep learning methods  and so on \cite{ho2019forecasting,cheng2021distance, HUANG2021106669}.
	
	In order to improve the prediction effect of CCI, this paper combines the ideas of information fusion and machine learning. Information fusion is a technology to fuse information from different sources to synthesize target data \cite{Xiao2020evidencecombination,Xiao2021GIQ,deng2021combining}. It is often used for intelligent decision-making, time series analysis and so on. This paper proposes a Multi-feature Fusion (MFF) neural network to predict CCI. 
	
	MFF uses the idea of pattern recognition to process time series in different feature. And MFF module generates a CCI feature sequence through the proposed sliding window and function sequence. The feature sequence saves the feature information of the CCI slices, and fuses the feature information into the required prediction data. Multi-layer perceptron here replaces the traditional information fusion method, which further improves the prediction effect. MFF neural network is composed of MFF module and Multi-layer perceptron. Experiments have proved that MFF has predictive accuracy in predicting CCI data. At the same time, MFF has the potential to further improve the accuracy of forecasting, and the proposal of MFF has made a contribution to time series forecasting.
	
	The structure of this paper is as follows: the second section introduces some basic theories of MFF, the third section is the definition of MFF, the fourth section shows the effect of predicting CCI and the analysis of CCI prediction, and the fifth section summarizes the paper.
	
	\section{Preliminaries}
	This section includes the basic theory of MFF. It supposes that the time series $T$ is as follows.
	$$T = \left\{(t_{1},v_{1}),(t_{2},v_{2}),(t_{3},v_{3}),(t_{4},v_{4}),...,(t_{n},v_{n})\right\} \eqno(1)$$
	where $t_i$ is the point in time, and $v_i$ is the value at point $t_i$ in the time series.
	
	The time series are treated as raw data as shown in Fig.1. The length of the time series $T$ is $n$.
	\begin{figure}[htbp]
		\centerline{\includegraphics[scale=0.6]{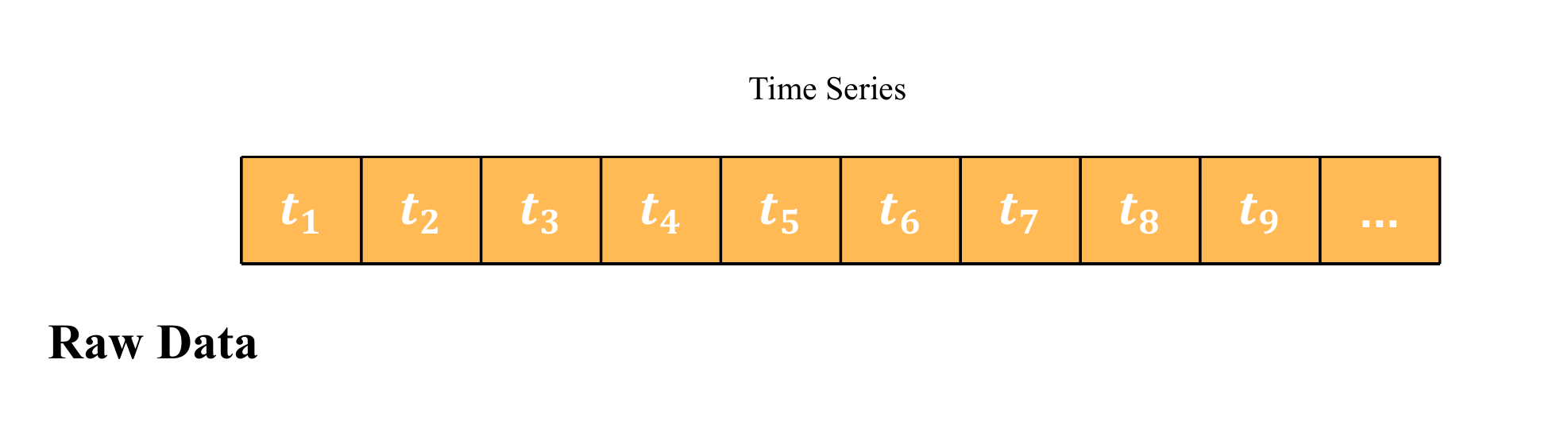}}
		\caption{The example of the raw data}
	\end{figure}

	\subsection{Sliding window and time slice set}
	Sliding window is a method in machine learning. By setting a fixed window size, data can be sliced by sliding. Assuming that the window size $Ws$ is a fixed integer ($Ws\leq n$, here $Ws=3$ is taken as an example), the process of sliding the window is shown in Fig.2.
	\begin{figure}[htbp]
		\centerline{\includegraphics[scale=0.45]{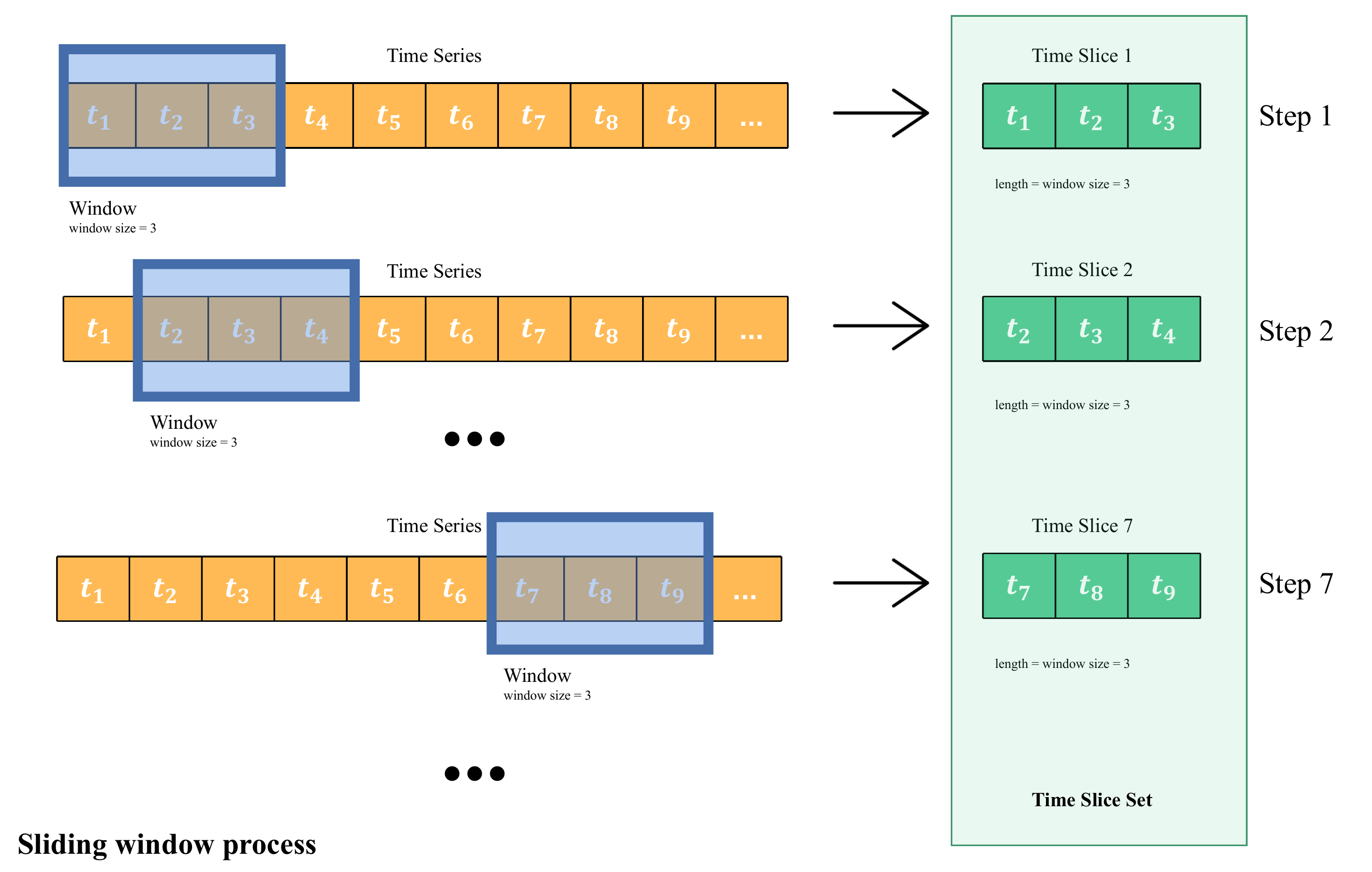}}
		\caption{The process of sliding window}
	\end{figure}
	
	\begin{myDef}
	The definition of the Sliding Window is as follows:
	$$SlidingWindow(T,Ws)=\left\{(t_i,TimeSlice_i)| 1\leqslant i \leqslant (n-Ws+1)\right\} \eqno(2)$$
	where time slice means a continuous subsequence of the original time series and the definition of the Time Slice is as follows:
	$$TimeSlice_i = \left\{v_i,v_{i+1},...,v_{i+Ws-1}\right\} \eqno(3)$$
	$$(t_i, v_i)\subseteq T \eqno(4)$$
	\end{myDef}

	\begin{myDef}
	The time slice set is the union of time slices generated by the time series through the sliding window as shown in Fig.3.
	$$TimeSliceSet=SlidingWindow(T,Ws) \eqno(5)$$
	\end{myDef}

	\begin{figure}[htbp]
		\centerline{\includegraphics[scale=0.55]{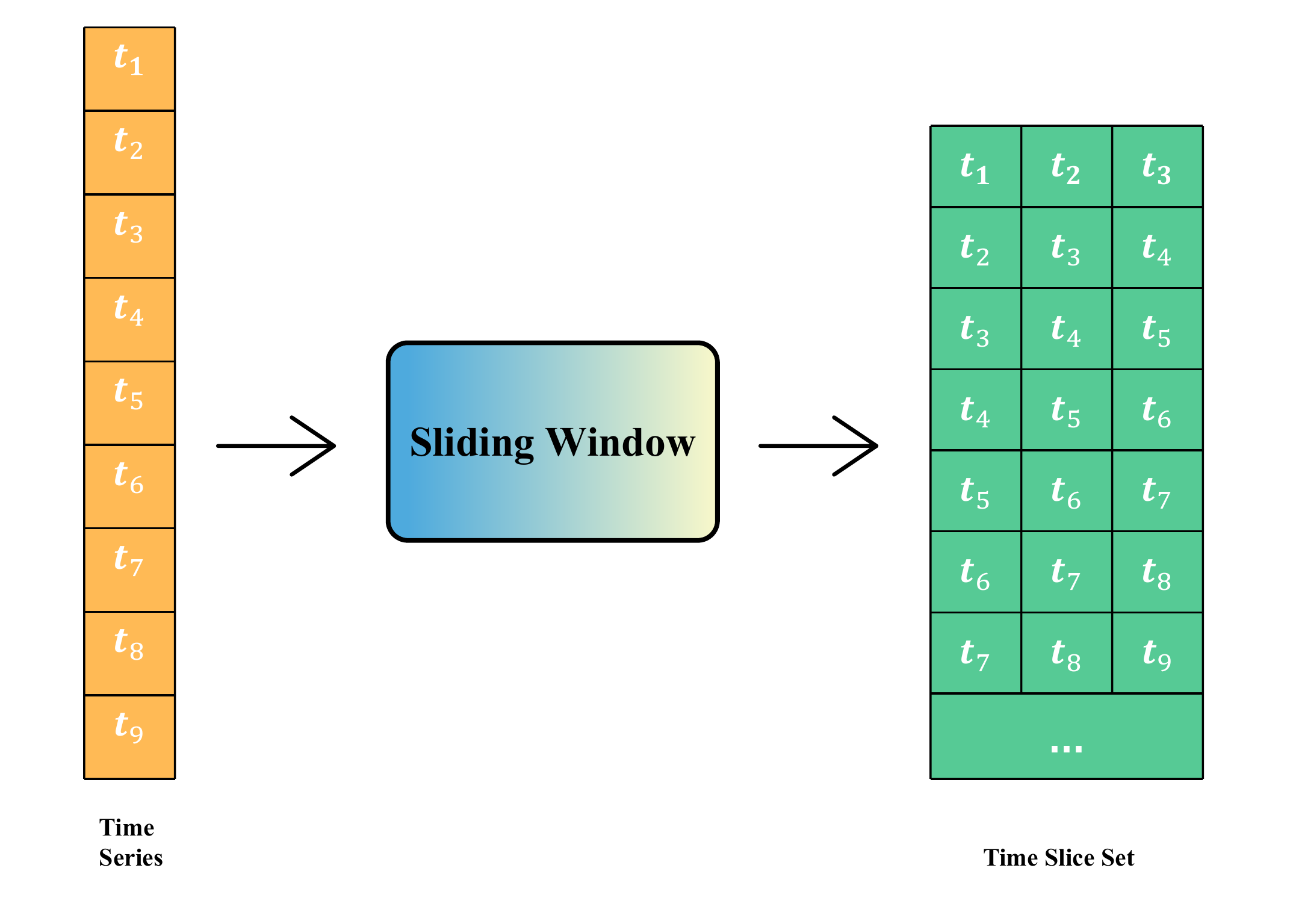}}
		\caption{The generation of time slice set }
	\end{figure}
	\subsection{Multilayer perceptron}
	The multi-layer perceptron (MLP) is promoted from the rerceptron learning algorithm (PLA) \cite{tang2015extreme}. Multilayer perceptron can effectively enhance the robustness of machine learning and the problem of overfitting. The structure of MLP is shown in Fig.4 below. Each node in the MLP sums the input according to the weight and bias, and the weight and bias will change during the optimization process.
	\begin{figure}[htbp]
		\centerline{\includegraphics[scale=0.65]{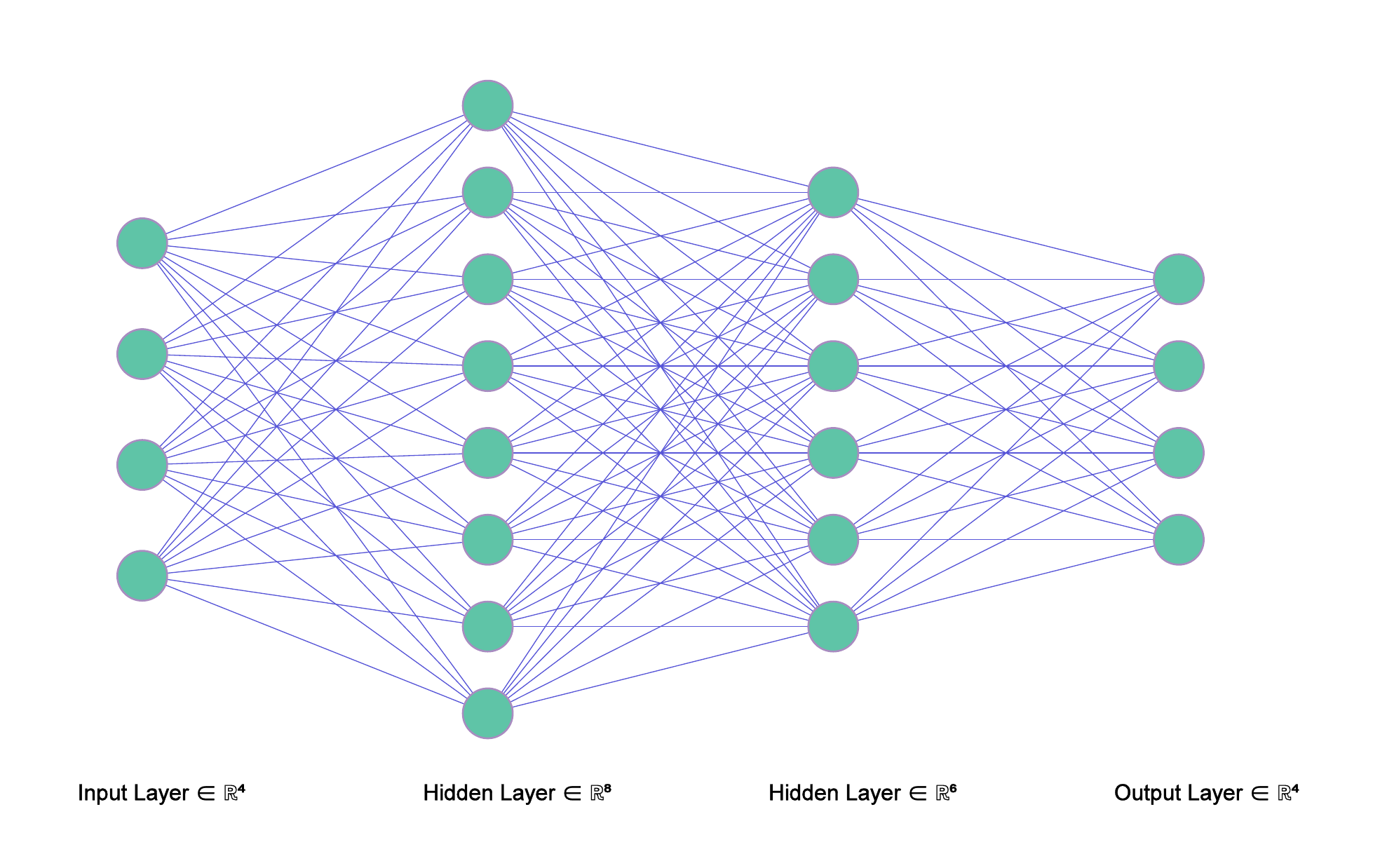}}
		\caption{The structure of MLP}
	\end{figure}
	
	\subsection{Mean squared error loss function}
	 Mean squared error (MSE) loss function is a loss function in machine learning \cite{pavlov2021using}. The mean squared error is defined as follows.
	 $$MSE_{Loss}=mean(L) \eqno(6)$$
	 $$L=l(x,y)=\left\{l_1,l_2,l_3,...,l_n\right\} \eqno(7)$$
	 $$l_i=(x_i-y_i)^2 \eqno(8)$$
	 where $x$ is the input, $y$ is the target, and the shapes of $x$ and $y$ are the same.
	 
	\subsection{Adam method} 
	Adam is an algorithm for first-order gradient-based optimization of stochastic objective functions, based on adaptive estimates of lower-order moments \cite{kingma2015adam,loshchilov2017decoupled}. Adam is simple to implement, has high computational efficiency, low memory requirements, and reduces the angle of the angle line, making it ideal for data and parameter problems \cite{kingma2015adam,loshchilov2017decoupled}. The pseudo code of Adam is as follows \cite{kingma2015adam,loshchilov2017decoupled}. And the good default parameters of Adam are shown in Tab.1 \cite{kingma2015adam,loshchilov2017decoupled}.
	
	\begin{algorithm}[htbp]
		\caption{Adam method}
		\begin{algorithmic}
			\Require $\alpha$:Stepsize
			\Require $\beta_1,\beta_2\in$[0,1): Exponential decay rates for the moment estimates
			\Require $f(\theta)$: Stochastic objective function with parameters $\theta$
			\Require: 
			\\ $\theta_0$: Initial parameter vector
			\\ $m_0\leftarrow 0$
			\\ $v_0\leftarrow 0$
			\\ $t\leftarrow 0$
			\While{$\theta_t$ not converged}
			\State $t \leftarrow t+1$
			\State $g_t\leftarrow\bigtriangledown_\theta f_t(\theta_{t-1})$
			\State $m_t\leftarrow\beta_1\cdot m_{t-1}+(1-\beta_1)\cdot g_t$
			\State $v_t\leftarrow\beta_2\cdot V_{t-1}+(1-\beta_2)\cdot g_t^2$
			\State $\hat{m_t}\leftarrow m_t/(1-\beta_1^t)$
			\State $\hat{v_t}\leftarrow v_t/(1-\beta_2^t)$
			\State $\theta_t\leftarrow\theta_{t-1}-\alpha\cdot\hat{m_t}/(\sqrt{\hat{v_t}}+\epsilon)$
			\EndWhile
			\Return $\theta_t$	
		\end{algorithmic}
	\end{algorithm}
	
	\begin{table}[htbp]
					\centering
		\setlength{\tabcolsep}{0.1mm}
		\begin{tabular}{|c|c|c|c|}
			\hline
			Parameter & Meaning                                                      & Good default settings         \\\hline
		$\alpha$	& Step Size                                                    & $0.001$                         \\ 
		$(\beta_1,\beta_2)$	& Exponential decay rates for the moment estimates             & $(0.9,0.999)$                 \\
		$\epsilon$	& Term added to the denominator                                 & $10^{-8}$                        \\ 
		$f(\theta)$	& Stochastic objective function with parameters  $\theta$              & \textbackslash{}            \\ \hline
		\end{tabular}
	\caption{The meaning and good default settings of Adam parameter}
	\end{table}
	
	\subsection{Cyclical learning rates} 
	Cyclical learning rates (CLR) is a method of dynamically adjusting the learning rate in machine learning. CLR eliminates the need for experiments to find the best value and timetable for the global learning rate. CLR does not reduce the learning rate in a monotonous manner, but rather makes the learning rate fluctuate between reasonable boundary values on a regular basis. The parameters and schematic diagram of CLR are shown in Tab.2 and Fig.5.
	
	\begin{table}[htbp]
					\centering
		\setlength{\tabcolsep}{0.1mm}
		\begin{tabular}{|c|c|}
			\hline
			Parameter          & Meaning                                                                                 \\\hline
			Base learning rate & Lower learning rate boundaries in the cycle for each parameter group \\
			Max learning rate  & Upper learning rate boundaries in the cycle for each parameter group                    \\
			Step size up       & Number of training iterations in the increasing half of a cycle                         \\
			Step size down     & Number of training iterations in the decreasing half of a cycle                        \\\hline
		\end{tabular}
	\caption{The meaning of CLR}
	\end{table}
	
	\begin{figure}[htbp]
		\centerline{\includegraphics[scale=0.5]{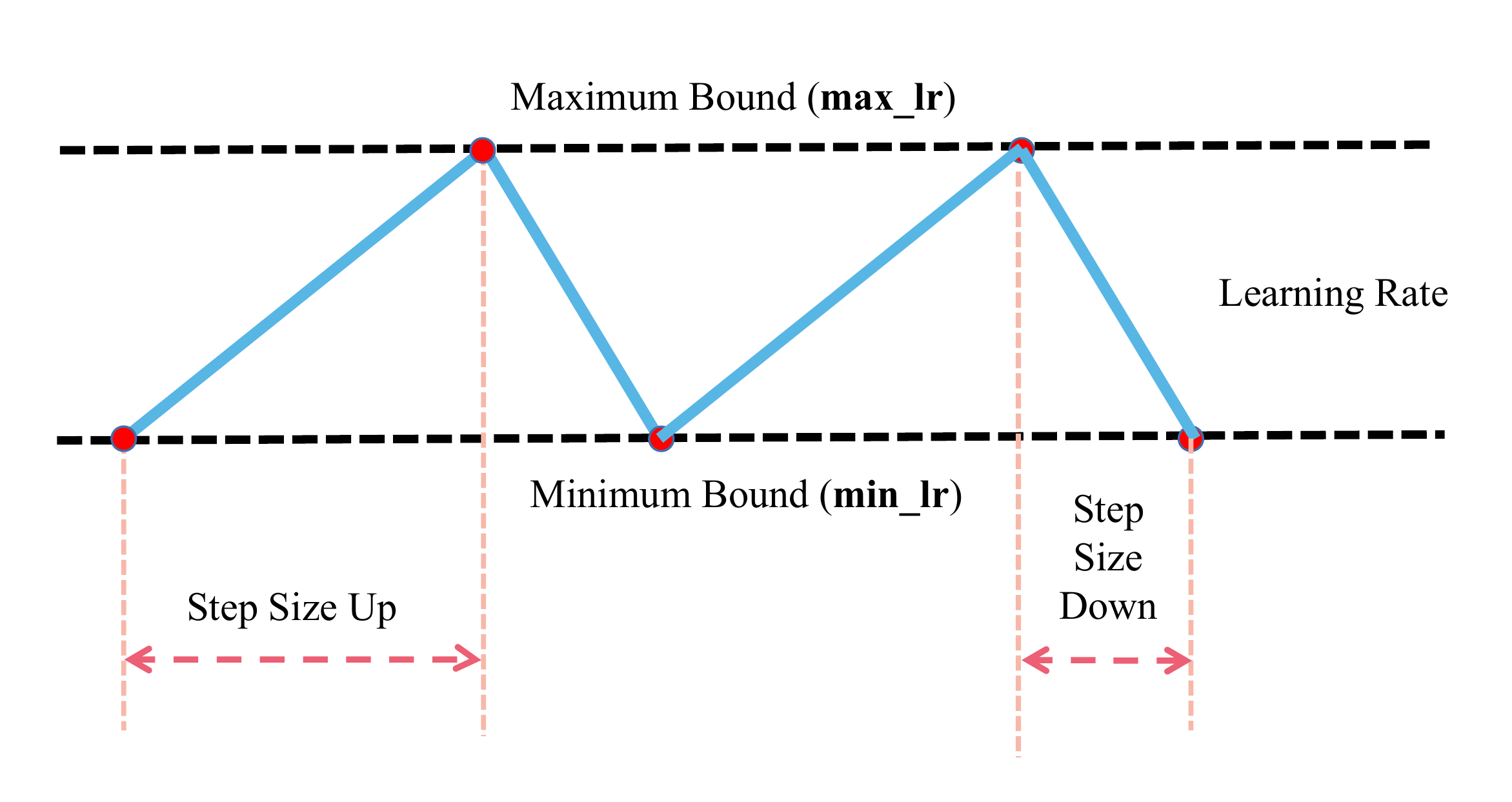}}
		\caption{Schematic diagram of CLR}
	\end{figure}

	\section{Multi-feature Fusion}
	\subsection{Step 1: Input time series}
	The input of MFF is the time series $T$. The time series $T$ is as follows:
	$$T = \left\{(t_{1},v_{1}),(t_{2},v_{2}),(t_{3},v_{3}),(t_{4},v_{4}),...,(t_{n},v_{n})\right\} \eqno(9)$$
	where $t_i$ is used as an index and does not exist in the form of $(t_i,v_i)$ tuples.
	
	\subsection{Step 2: Slice time series}
	When generating a time slice set, MFF needs to determine the size of a sliding window $Ws$. The calculation process of Time slice set $S_T$ is as follows:
	$$S_T=SlidingWindow(T,Ws) \eqno(10)$$
	
	When generating a time slice set, the setting of Ws needs to be considered. The number of time slices is $(n-Ws)$. Excessive $Ws$ results in fewer slices and fewer learning samples. If $Ws$ is too small, each sample can only reflect short time series characteristics. $Ws\approx\frac{1}{2}n$ is default parameters. The shape of $S_T$ is $((n-Ws+1),Ws)$.
	\subsection{Step 3: Input function sequence}
	In step 3, MFF needs to complete the preprocessing of the time slice set $S_T$ and convert the time slice into a feature sequence. The function sequence is a converter that converts the time slice into a feature sequence as shown in Fig.6.
	
	\begin{figure}[htbp]
		\centerline{\includegraphics[scale=0.6]{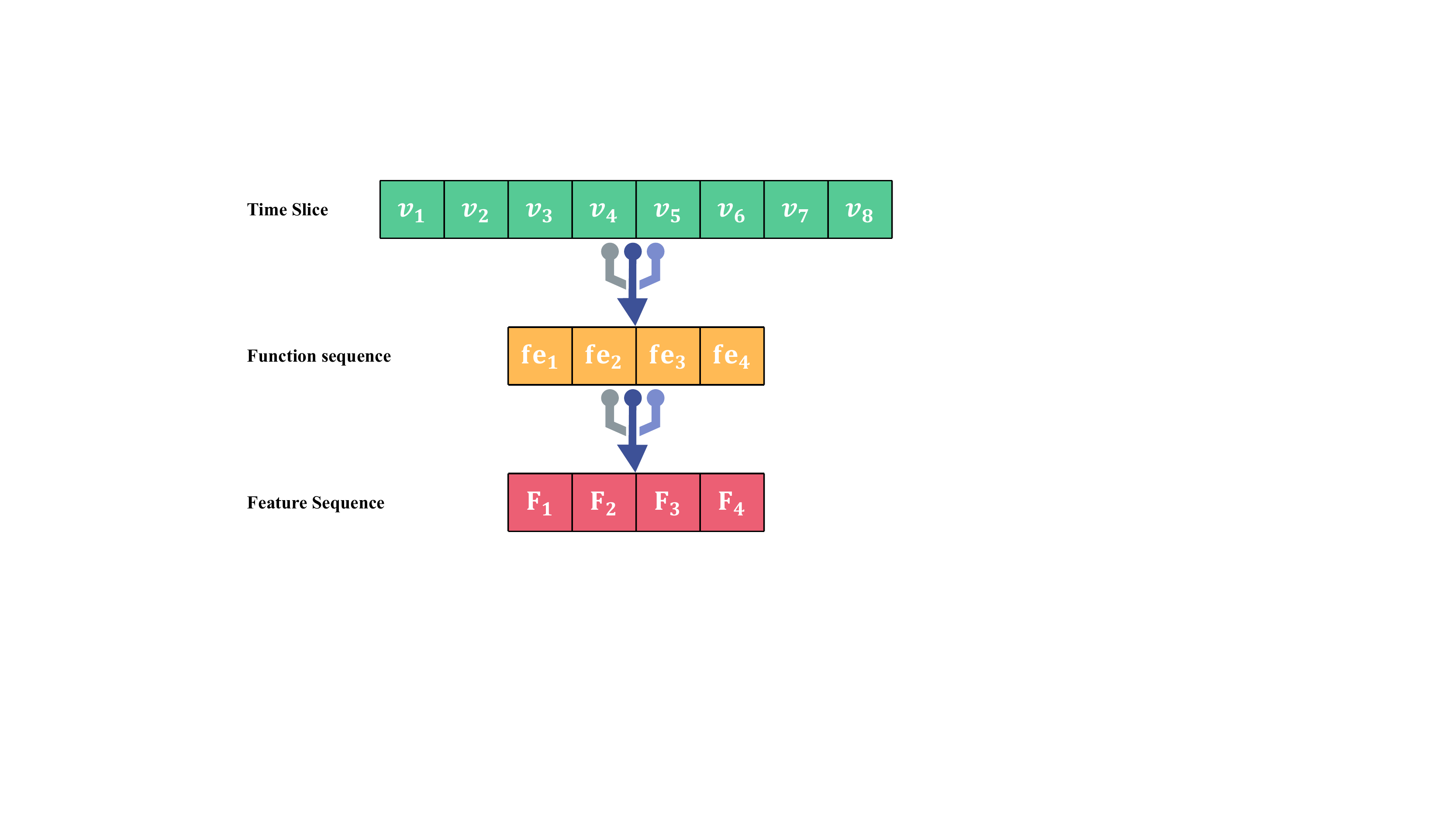}}
		\caption{Example of feature conversion (Window size=8, there are four functions in the function sequence. $v_i$ is the value corresponding to the time node $i$ in the time series.)}
	\end{figure}

	\begin{myDef}
	Function sequence is a set of functions, defined as follows:
	$$Fs(x)=\left\{F_1(x), F_2(x), ...,F_m(x)\right\} \eqno(11)$$
	$$F_i(x) = f_{e\ i} \eqno(12)$$
	where $m$ is the number of functions in the function sequence $Fs$, $x$ is a time slice and $F_i(x)$ transfers $x$ which is in the shape $(1\times Ws)$ to feature $f_{ei}$ which is in the shape of $(1\times 1)$.
	\end{myDef}
	
	 After the function sequence is input, the time slice set $S_T$ is converted to the feature sequence set $S_{F}$ as follows:
	$$S_F=Fs(S_T)=\left\{Fs(S_{T\ 1}), Fs(S_{T\ 2}), ...,Fs(S_{T\ n-Ws+1})\right\} \eqno(13)$$
	$$Fs(S_{T\ i})=\left\{f_{e(i,1)},f_{e(i,2)},...,f_{e(i,m)}\right\} \eqno(14)$$
	
	$f_{e(i,j)}$ represents the feature value generated by the function $F_j(x)$ in the time slice $S_{T\ i}$. For different training targets and training data, the feature function $F(x)$ selected in MFF for prediction is different. The shape of feature sequence set $S_F$ is $((n-Ws+1),m)$. The MFF module is composed of sliding window and function sequence processing.

	\subsection{Step 4: Multilayer perceptron: forward propagation}
	In MFF, MLP has four layers: input layer, hidden layer 1, hidden layer 2 and output layer. The nodes in the three layers are $m$, $n_1$, $n_2$ and 1 as shown in Fig.7. Each feature sequence will be input into MLP, and then a result will be input. Whenever the result corresponding to the feature sequence is generated, it will do back propagate and optimize the parameters. A forward propagation and back propagation are called an epoch. Each epoch will update the result of the result as follows:
	$$result\leftarrow MLP(m,n_1,n_2) \eqno(15)$$
	
	\begin{figure}[htbp]
		\centerline{\includegraphics[scale=0.45]{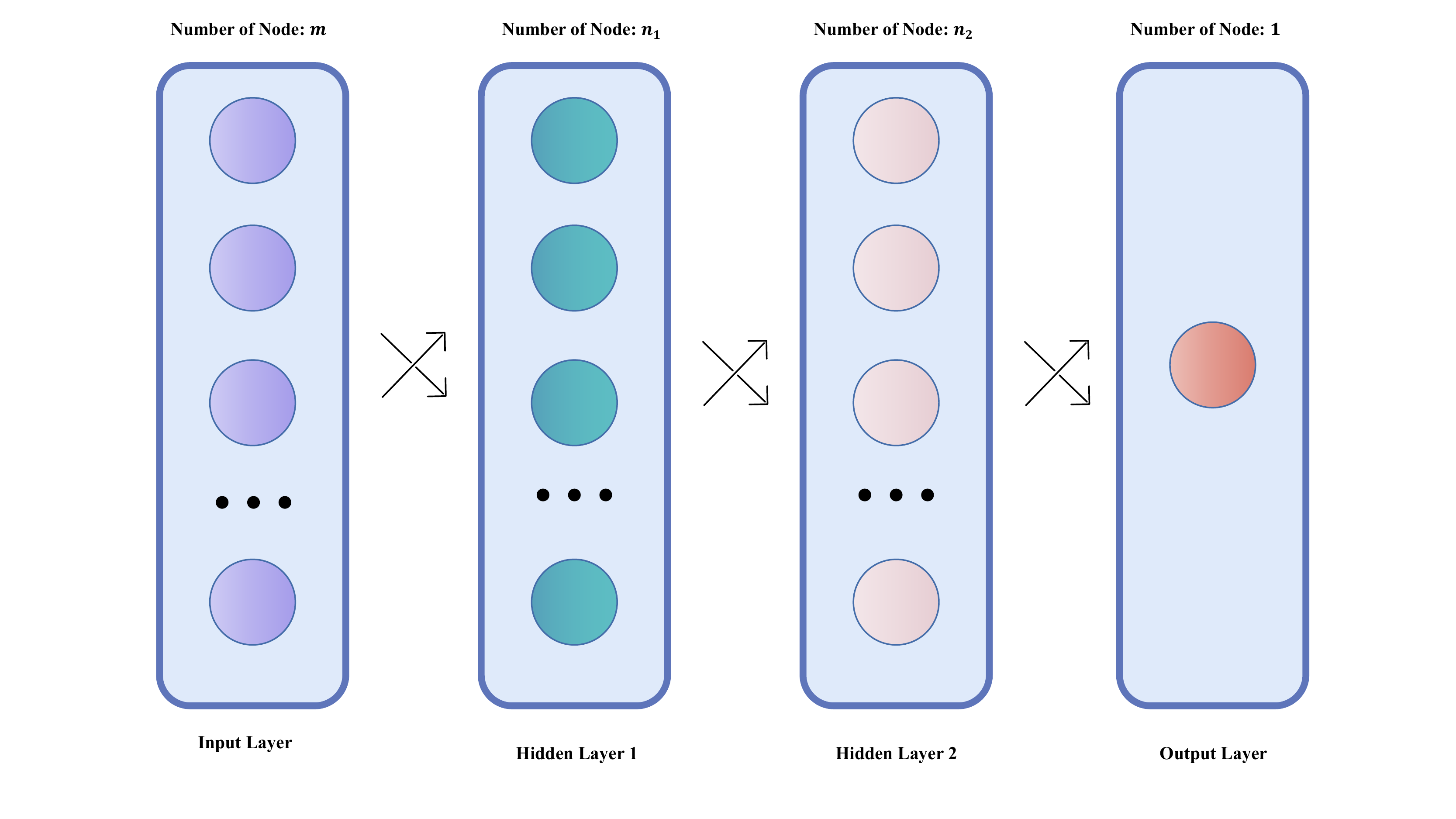}}
		\caption{The structure of MLP in the MFF}
	\end{figure}

	\subsection{Step 5: Multilayer perceptron: back propagation and parameter optimization}
	In MFF, each epoch needs back propagation and parameter optimization. The loss function of MFF is MSE and the target is next time node's value of the the current time slice. After calculating the loss in each epoch, the parameters of MFF are back-propagated and optimized by Adam algorithm and CLR. When initializing MFF, it is necessary to input the upper and lower limits of the learning rate, which are dynamically adjusted by the CLR algorithm during training. MFF does not use the traditional gradient descent method of MLP, but uses the Adam algorithm for gradient descent, which accelerates machine learning and strengthens the effect of machine learning. 
	
	The loss and model parameters calculated in each epoch will be saved in a set. In MFF, the number of epochs $N$ is a variable set in advance. After the back propagation and parameter optimization of each epoch updated by the Adam algorithm and CLR, MFF returns to Step 4 for the next epoch training. When the last epoch is completed, a set of training parameters with the smallest loss will be selected for prediction. The process of MFF is shown in Fig.8 and the pseudo code of MFF is as follows.
	
	\subsection{Step 6: Predict}
	In MFF, the model parameter with the smallest loss is applied to the MLP and then the time series that needs to be predicted are input into the MFF to complete the prediction.
	
	\begin{figure}[htbp]
		\centerline{\includegraphics[scale=0.55]{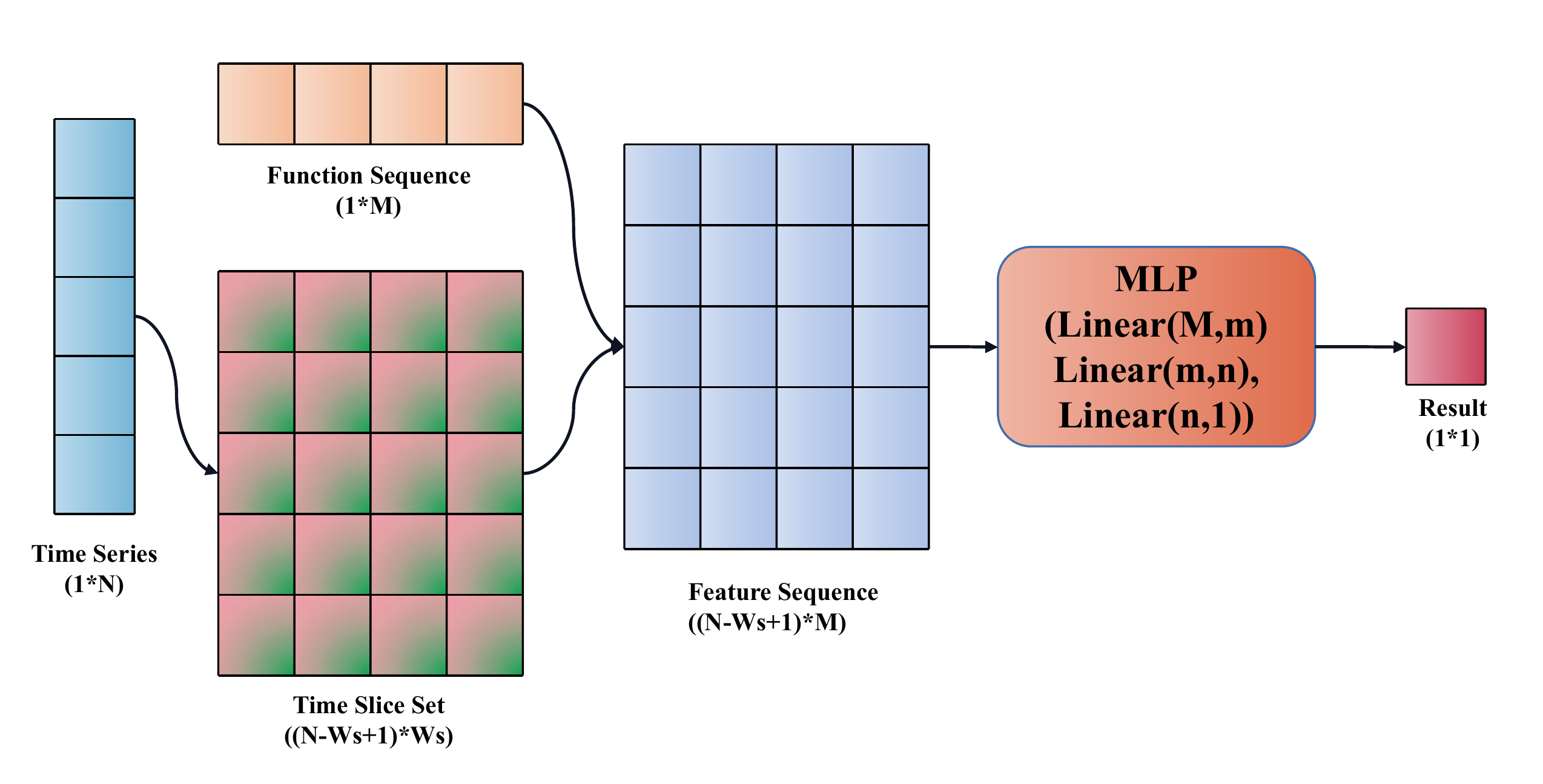}}
		\caption{The process of MFF}
	\end{figure}

	\begin{figure}[htbp]
		\centerline{\includegraphics[scale=0.15]{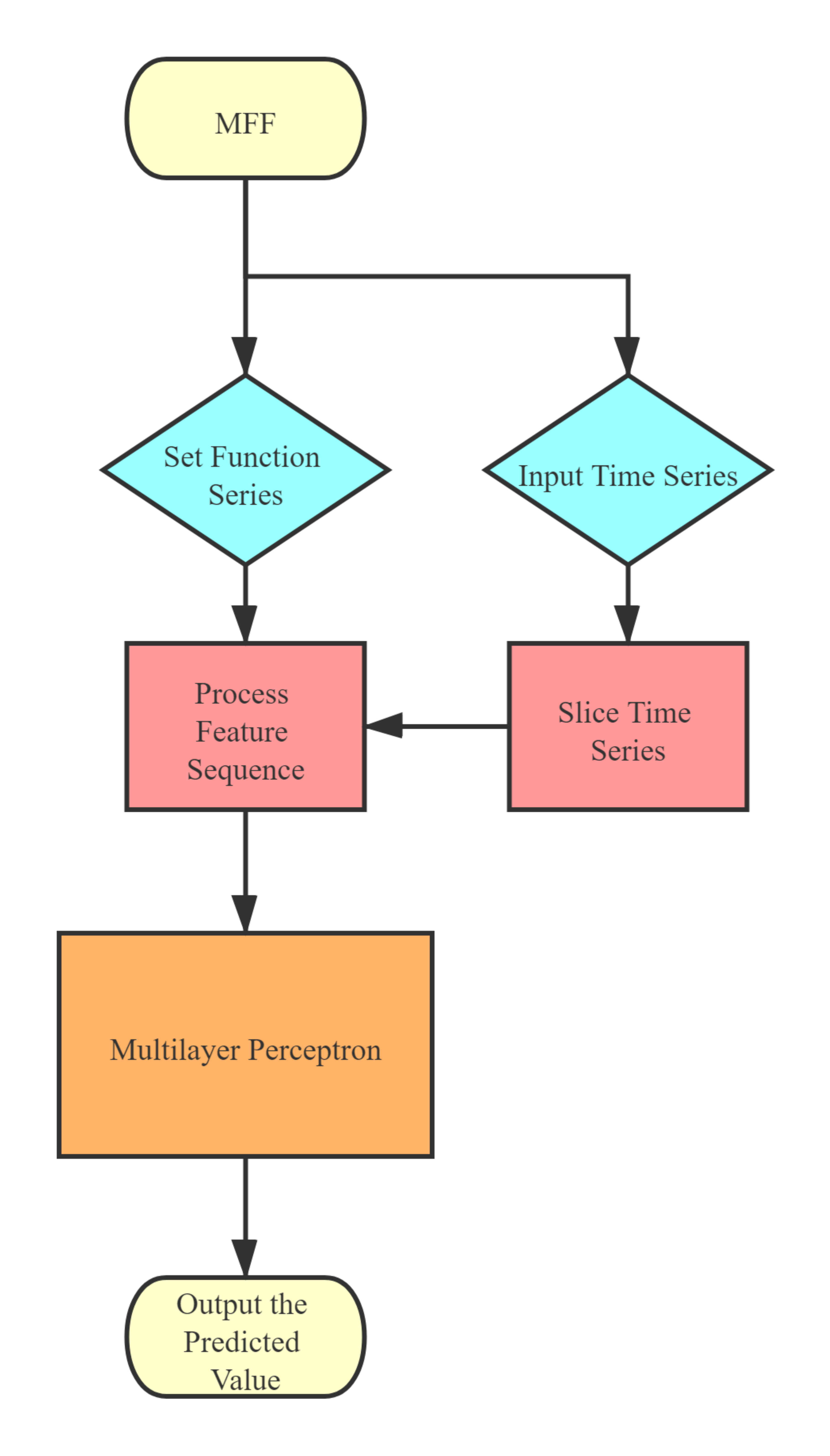}}
		\caption{The flow chart of MFF}
	\end{figure}
	
	\begin{algorithm}[htbp]
		\caption{$MFF(T,Ws,F_s,N, Shape)$}
		\begin{algorithmic}[1]
			\Require Time series $T$ 
			\Require Sliding window size $Ws$
			\Require Function Sequence $F_s$
			\Require Number of epoch $N$ 
			\Require Shape of MLP 
			\State Slice time series by sliding window algorithm
			\State Generate feature sequence $S_F$ hrough time slice set $S_T$ and function sequence $F_s$
			\State Train: Set $F_s$ as training set
			\For {Epoch = 1 to $N$ }
				\State MLP: forward propagation
				\State MLP: back propagation and parameter optimization by Adam and CLR algorithm
				\State Save model parameter and loss in model set $S_M$
			\EndFor
			\State Predict: Apply the model with minimum loss in the MLP
			\State Input last feature and output the result $\hat{y}_{n+1}$	\\
			\Return $\hat{y}_{n+1}$	
		\end{algorithmic}
	\end{algorithm}
	
	\section{Experiment}
	\subsection{Data set description}
	Engineering News Record (ENR) is a monthly publication that publishes the CCI \cite{shahandashti2013forecasting,hwang2011time}. CCI has been studied by many civil engineers and cost analysts because it contains vital building industry price information. The CCI data set includes a total of 295 data values of construction costs from January 1990 to July 2014.
	
	\subsection{Experiment preprocessing}
	For CCI, MFF needs to determine the size of a window, $Ws=180$ in the experiment as an example. At the same time, the last data is used as the target of the penultimate time point, without sliding window. A total of 116 time slices were generated, and there were 116 corresponding feature sequences. In this experiment, 116 pieces of data are divided into experimental set and test set according to the ratio of $8:2$.
	
	The choice of function is variable. In this experiment, the MFF function sequence is composed of 6 functions. The function names and definitions are shown in Tab.3. Also, the number of nodes $(m, n)$ of MLP is set to $(8, 5)$  and the max epoch is 10000 in this experiment. In the CLR algorithm, the base learning rate is $10^{-12}$ and the max learning rate is $10^{-4}$. In the MFF training process, the gradient descent uses the Adam algorithm, which can improve the training efficiency of the model.
	
	\begin{table}[htbp]
					\centering
		\begin{tabular}{cc}
			\hline
			Function           & Definition                                                        \\ \hline
			Index              & The order of time nodes in the current slice                      \\
			Mean               & Average of the time series                                        \\
			Standard deviation & Standard deviation of the time series                             \\
			Distance           & Time series maximum minus minimum                                 \\
			ApEn               & Approximate entropy of time series \cite{montesinos2018use}                                \\
			Degree             & The sum of the degrees of the visibility graph \cite{lacasa2008time} \\ \hline
		\end{tabular}
	\caption{Function sequence in the experiment   ($F(x)$ selected in this experiment)}
	\end{table}

	\subsection{Experimental results}
	To evaluate the prediction of each method, there are five measures of error: mean absolute difference (MAD) \cite{yitzhaki2003gini} , mean absolute percentage error (MAPE) \cite{de2016mean} , symmetric mean absolute percentage error (SMAPE) \cite{tofallis2015better}, root mean square error (RMSE) \cite{hyndman2006another} , and normalized root mean squared error (NRMSE) \cite{shcherbakov2013survey} :
	$$MAD=\frac{1}{N}\sum_{t=1}^{N}\left|\hat{y}(t)-y(t)\right| \eqno(16)$$
	$$MAPE=\frac{1}{N}\sum_{t=1}^{N}\frac{\left|\hat{y}(t)-y(t)\right|}{y(t)} \eqno(17)$$
	$$SMAPE=\frac{2}{N}\sum_{t=1}^{N}\frac{\left|\hat{y}(t)-y(t)\right|}{\hat{y}(t)+y(t)} \eqno(18)$$
	$$RMSE=\sqrt{\frac{1}{N}\sum_{t=1}^{N}\left|\hat{y}(t)-y(t)\right|^{2}} \eqno(19)$$
	$$NRMSE=\frac{\sqrt{\frac{1}{N}\sum_{t=1}^{N}\left|\hat{y}(t)-y(t)\right|^{2}}}{y_{max}-y_{min}} \eqno(20)$$
	where $\hat y(t)$ is the predicted value, $y(t)$ is the true value and N is the total number of $\hat y(t)$.
	
	Fig.10 shows the prediction of MFF$(8,5)$. The predicted value of MFF is close to the actual value, and the prediction effect is good.

	\begin{figure}[htbp]
		\centerline{\includegraphics[scale=0.8]{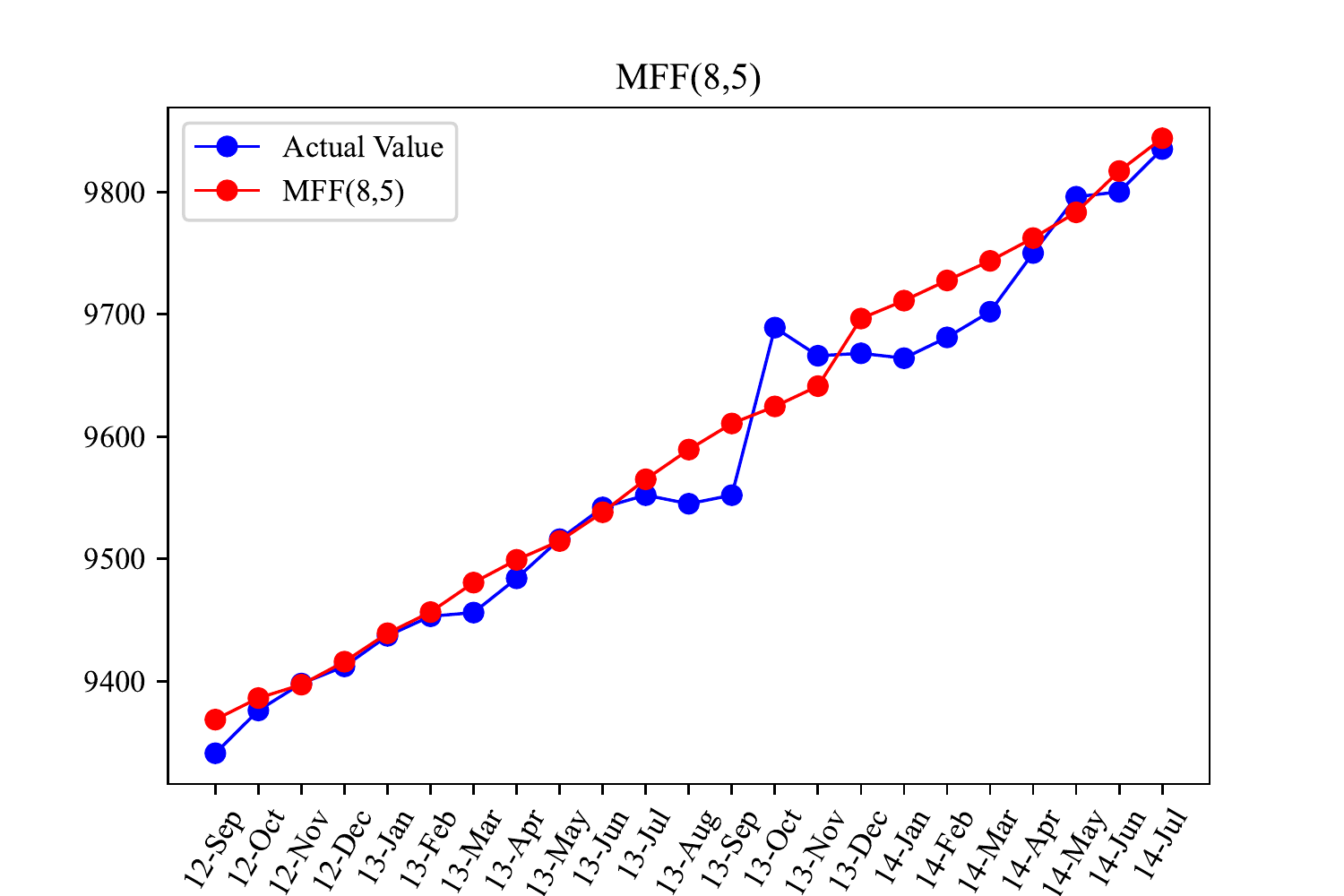}}
		\caption{Prediction of MFF$(8,5)$}
	\end{figure}

	\subsection{Comparative Experiment}
	In order to verify the prediction effect of MFF, the prediction results of MFF will be compared with three different types of prediction methods statistical prediction methods, machine learning regression methods and hybrid prediction methods. At the same time, in order to distinguish it from the existing deep learning methods, MFF and MLP, Convolutional Neural Network (CNN) and Long-Short Term Memory (LSTM) prediction methods have been ablated experiments.
	
	\subsubsection{Comparison between MFF and statistical prediction methods}
	
		In this section, MFF will be compared with statistical prediction methods. Among the statistical comparison methods, Simple Moving Average (SMA) (K=1) \cite{guan2017two}, Autoregressive Integrated Moving Average model (ARIMA) \cite{tseng2002combining}, Seasonal Autoregressive Integrated Moving Average model (Seasonal ARIMA) \cite{tseng2002fuzzy} and ExponenTial Smoothing (ETS) \cite{billah2006exponential} are commonly used methods for prediction. Random walk is also a commonly used prediction method in statistics. Mao and Xiao proposed a random walk prediction method based on complex networks, which has good prediction performance and will also be used as a comparison method \cite{mao2019time}. 
		
		Tab.4 and Fig.11 are the comparison of the experimental effects of MFF and statistical prediction methods. According to the experimental results, MFF performs better than the statistical methods mentioned above. In Fig.11, the statistical method predicts the result is relatively stable, the trend is similar to the true value, but compared with MFF, MFF is more stable, and MFF is closer to the actual value.

		\begin{table}[htbp]
						\centering
		\begin{tabular}{cccccc}
			\hline
			& MAD              & MAPE            & SMAPE           & RMSE             & NRMSE             \\ \hline
			ETS  \cite{billah2006exponential} & 55.6591          & 0.5838          & 0.5816          & 64.4560          & 300.9969          \\
			Seasonal ARIMA \cite{tseng2002fuzzy}       & 45.3349          & 0.4769          & 0.4753          & 54.8709          & 240.0670          \\
			SMA(K=1) \cite{guan2017two}             & 43.7391          & 0.4582          & 0.4566          & 55.8180          & 256.9233          \\
			ARIMA  \cite{tseng2002combining}               & 38.6931          & 0.4055          & 0.4044          & 47.7177          & 214.7822          \\
			Mao and Xiao's Method \cite{mao2019time} & 37.7301          & 0.3940          & 0.3928          & 49.7481          & 226.0986          \\
			\textbf{MFF(8,5)}     & \textbf{22.2877} & \textbf{0.2318} & \textbf{0.2316} & \textbf{29.2458} & \textbf{131.5833} \\ \hline
		\end{tabular}
		\caption{Forecasting error of MFF and statistical prediction methods}
	\end{table}

	\begin{figure}[htbp]
		\centerline{\includegraphics[scale=0.8]{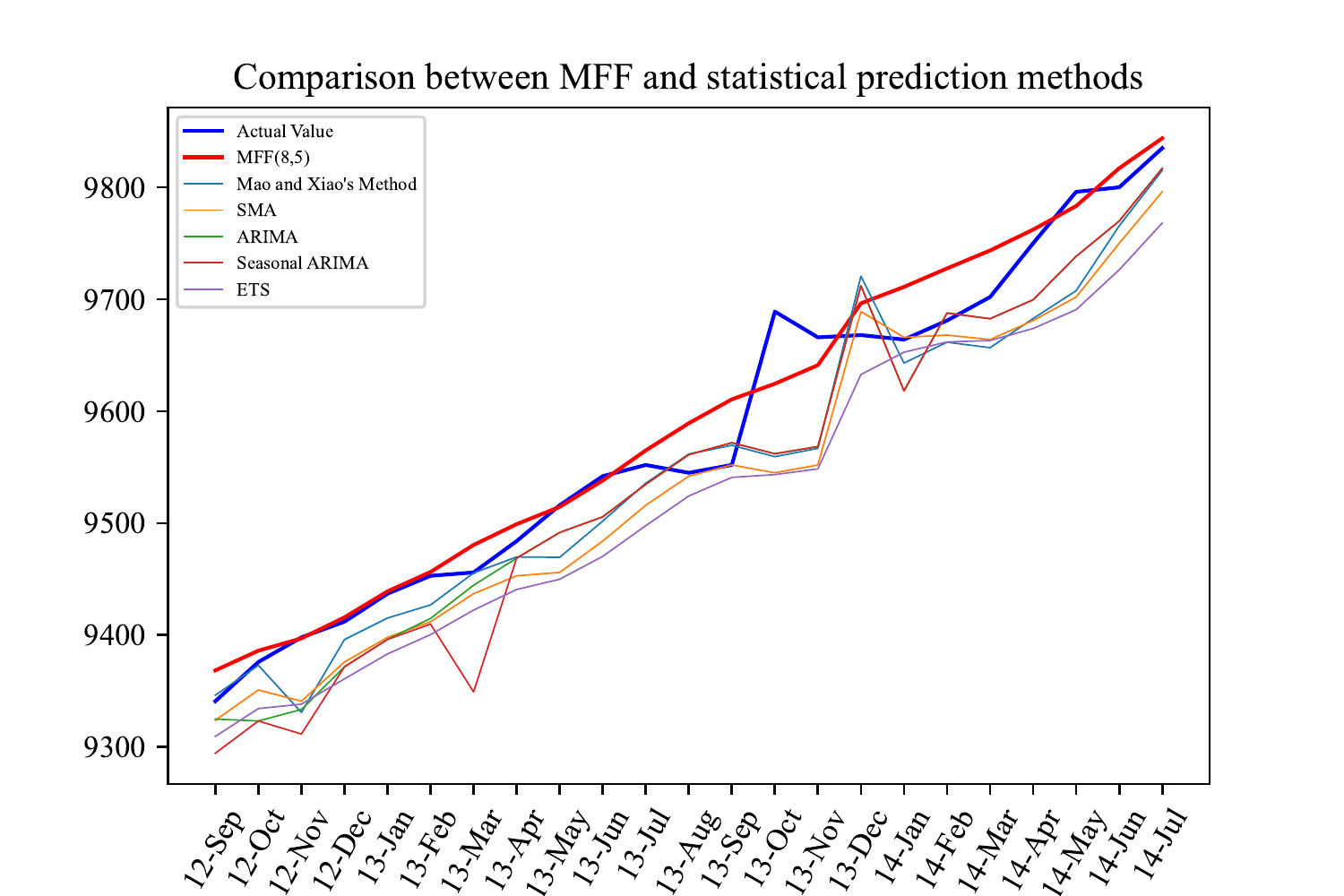}}
		\caption{Comparison between MFF$(8,5)$ and statistical prediction methods}
	\end{figure}
	
	\subsubsection{Comparison between MFF and machine learning regression methods}
	
			In this section, MFF will be compared with machine learning regression methods. Among the machine learning comparison methods,  Decision Tree Regression (DTR) \cite{hastie2009elements}, Ordinary least squares Linear Regression (Linear) \cite{hutcheson2011ordinary}, Lasso model fit with Least Angle Regression (Lasso) \cite{taylor2014post}, Support Vector Machines Regression (SVM) \cite{tong2009analysis}, Bayesian Ridge Regression (Bayesian) \cite{xu2020blood} and Logistic Regression (Logistic) \cite{hosmer2013applied,defazio2014saga} are commonly used methods for prediction. 
	
			Tab.5 and Fig.12 are the comparison of the experimental effects of MFF and machine learning regression methods. According to the experimental results, MFF performs better than the machine learning methods mentioned above. The trend of machine learning regression methods is stable and more accurate than statistical methods. There is a gap between the predicted value and MFF, and MFF is closer to the true value. At the same time, the jitter of MFF is small, and the trend is close to the real trend of CCI.

	\begin{figure}[htbp]
		\centerline{\includegraphics[scale=0.8]{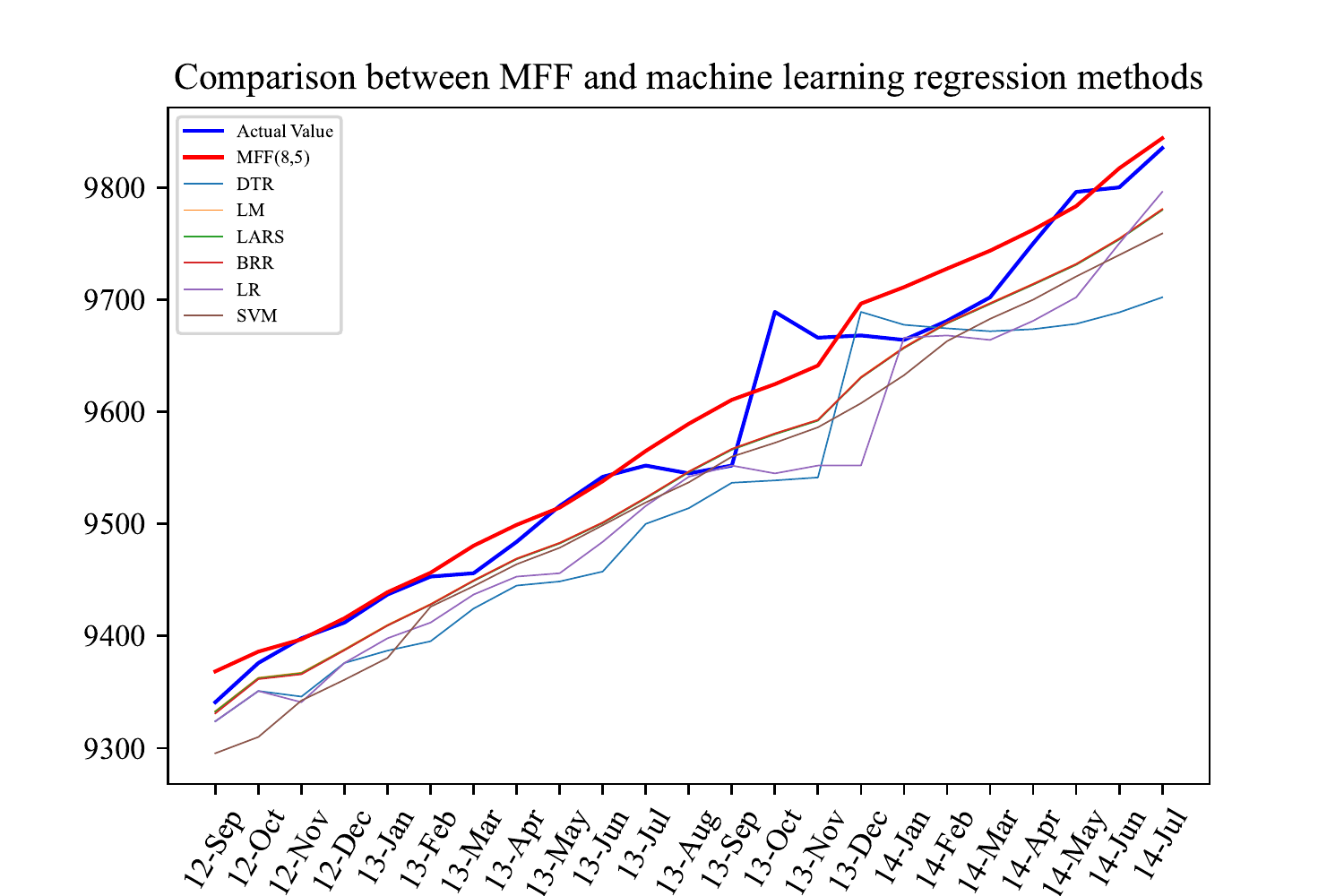}}
		\caption{Comparison between MFF$(8,5)$ and other methods}
	\end{figure}

		\begin{table}[htbp]
			\centering
		\begin{tabular}{cccccc}
			\hline
			& MAD              & MAPE            & SMAPE           & RMSE             & NRMSE             \\ \hline
			DTR \cite{hastie2009elements} & 58.3954          & 0.6117          & 0.6089          & 71.7173          & 368.8740          \\
			Linear \cite{hutcheson2011ordinary}                       & 47.8696          & 0.5016          & 0.4996          & 60.6755          & 279.2818          \\
			SVM  \cite{tong2009analysis}                   & 45.6480          & 0.4784          & 0.4769          & 52.7443          & 245.0101          \\
			Lasso \cite{taylor2014post}                    & 30.9693          & 0.3232          & 0.3224          & 40.1681          & 189.9494          \\
			Bayesian  \cite{xu2020blood}                    & 30.8234          & 0.3218          & 0.3209          & 39.8843          & 188.1513          \\
			Logistic  \cite{hosmer2013applied,defazio2014saga}                     & 30.3914          & 0.3172          & 0.3163          & 39.6220          & 187.2685          \\
			\textbf{MFF(8,5)}        & \textbf{22.2877} & \textbf{0.2318} & \textbf{0.2316} & \textbf{29.2458} & \textbf{131.5833} \\ \hline
		\end{tabular}
		\caption{Forecasting error of MFF and machine learning regression methods}
	\end{table}

	\subsubsection{Comparison between MFF and hybrid prediction methods}
	In this section, MFF will be compared with hybrid prediction methods. Hybrid model is a method for time series forecasting to improve forecast accuracy. The use of hybrid model can combine the linear characteristics of statistical methods and the characteristics of machine learning nonlinear prediction Artificial Neural Network (ANN) to further increase the accuracy of prediction. Common hybrid models are ARIMA-ANN \cite{babu2014moving} and ETS-ANN \cite{panigrahi2017hybrid}.
	
	This experiment takes a parallel approach in the experiment of the hybrid model. In order to ensure the rigor of the experiment, the ANN hybrid model and MFF in the hybrid model use the same training parameters include Adam algorithm and CLR.
	
	Tab.6 and Fig.13 are the comparison of the experimental effects of MFF and hybrid prediction methods. According to the experimental results, MFF performs better than the hybrid prediction methods mentioned above. The hybrid model further improves the accuracy of prediction. ARIMA and ETS as statistical models have good prediction performance. ANN as a non-linear model for processing ARIMA and ETS further improves the accuracy and reduces errors. But compared with MFF, the hybrid model has a greater degree of jitter.

	\begin{figure}[htbp]
		\centerline{\includegraphics[scale=0.8]{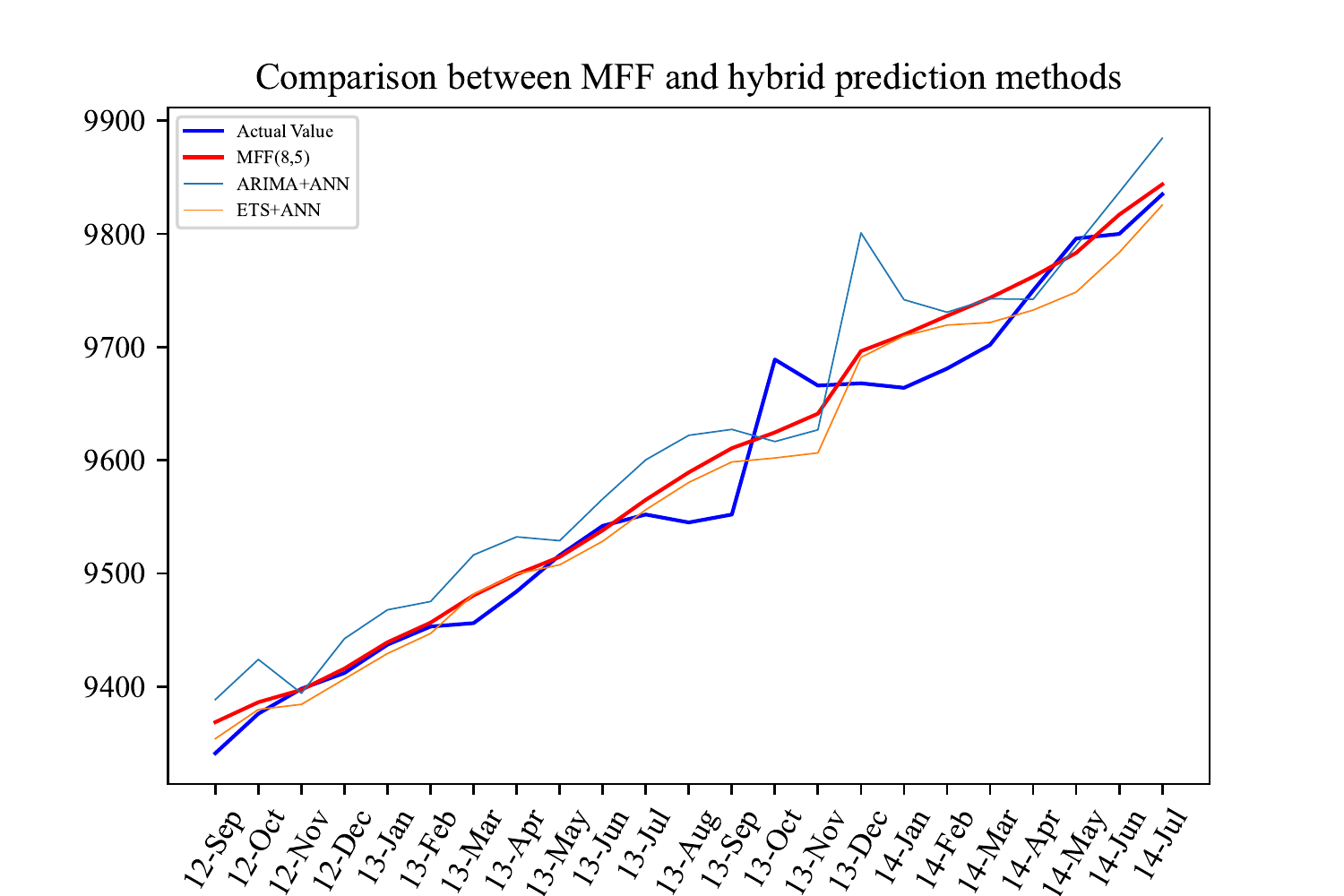}}
		\caption{Comparison of MFF$(8,5)$ and hybrid prediction methods}
	\end{figure}

	\begin{table}[htbp]
					\centering
		\begin{tabular}{cccccc}
			\hline
			& MAD              & MAPE            & SMAPE           & RMSE             &  NRMSE             \\ \hline
			ARIMA-ANN \cite{babu2014moving}        & 45.2997          & 0.4701          & 0.4713          & 53.6162          & 240.7094          \\
			ETS-ANN \cite{panigrahi2017hybrid}          & 24.4770          & 0.2545          & 0.2543          & 32.0290          & 147.4822          \\
			\textbf{MFF(8,5)} & \textbf{22.2877} & \textbf{0.2318} & \textbf{0.2316} & \textbf{29.2458} & \textbf{131.5833} \\ \hline
		\end{tabular}
		\caption{Forecasting error of MFF and hybrid prediction methods}
	\end{table}

	\subsubsection{Ablation experiment}
	
\begin{table}[htbp]
	\centering
	\setlength{\tabcolsep}{1mm}
	\begin{tabular}{cccccc}
		\hline
		& \begin{tabular}[c]{@{}c@{}}Structure\\ (Hidden Layer/ \\ Fully connected Layer/\\  Convolution Layer/ \\ Pooling Layer/\\ Dense Layer)\end{tabular} & \begin{tabular}[c]{@{}c@{}}Training \\ Iteration\end{tabular} & \begin{tabular}[c]{@{}c@{}}Learning \\ Rate\end{tabular} & \begin{tabular}[c]{@{}c@{}}Loss\\ Function\end{tabular} & \begin{tabular}[c]{@{}c@{}}Optimizer\\ Function\end{tabular} \\ \hline
		& Hidden Layer=100                                                                                                                                    &                                                               &                                                          &                                                         &                                                              \\
		MLP      & Activation=Relu                                                                                                                                     & 10000                                                         & 0.01                                                     & MSE                                                     & Adam                                                         \\
		& Output Layer=1                                                                                                                                      &                                                               &                                                          &                                                         &                                                              \\ \hline
		& Hidden Layer=50                                                                                                                                     &                                                               &                                                          &                                                         &                                                              \\
		& Activation=Relu                                                                                                                                     &                                                               &                                                          &                                                         &                                                              \\
		LSTM     & Input Timestep=3                                                                                                                                    & 10000                                                         & 0.01                                                     & MSE                                                     & Adam                                                         \\
		& Output Timestep=1                                                                                                                                   &                                                               &                                                          &                                                         &                                                              \\
		& Dense Layer=1                                                                                                                                       &                                                               &                                                          &                                                         &                                                              \\ \hline
		& Convolution Layer:                                                                                                                                  &                                                               &                                                          &                                                         &                                                              \\
		& Filters=64                                                                                                                                          &                                                               &                                                          &                                                         &                                                              \\
		& Kernel Size=2                                                                                                                                       &                                                               &                                                          &                                                         &                                                              \\
		CNN      & Activation=Relu                                                                                                                                     & 10000                                                         & 0.01                                                     & MSE                                                     & Adam                                                         \\
		& Pooling Layer:                                                                                                                                      &                                                               &                                                          &                                                         &                                                              \\
		& Pool Size=2                                                                                                                                         &                                                               &                                                          &                                                         &                                                              \\
		& Dense Layer=100                                                                                                                                     &                                                               &                                                          &                                                         &                                                              \\
		& Dense Layer=1                                                                                                                                       &                                                               &                                                          &                                                         &                                                              \\ \hline
		& MFF Layer                                                                                                                                           &                                                               &                                                          &                                                         &                                                              \\
		& Hidden Layer=50                                                                                                                                     &                                                               &                                                          &                                                         &                                                              \\
		MFF+LSTM & Activation=Relu                                                                                                                                     & 10000                                                         & 0.01                                                     & MSE                                                     & Adam                                                         \\
		& Input Timestep=3                                                                                                                                    &                                                               &                                                          &                                                         &                                                              \\
		& Output Timestep=1                                                                                                                                   &                                                               &                                                          &                                                         &                                                              \\
		& Dense Layer=1                                                                                                                                       &                                                               &                                                          &                                                         &                                                              \\ \hline
		& MFF Layer                                                                                                                                           &                                                               &                                                          &                                                         &                                                              \\
		& Convolution Layer:                                                                                                                                  &                                                               &                                                          &                                                         &                                                              \\
		& Filters=64                                                                                                                                          &                                                               &                                                          &                                                         &                                                              \\
		& Kernel Size=2                                                                                                                                       &                                                               &                                                          &                                                         &                                                              \\
		MFF+CNN  & Activation=Relu                                                                                                                                     & 10000                                                         & 0.01                                                     & MSE                                                     & Adam                                                         \\
		& Pooling Layer:                                                                                                                                      &                                                               &                                                          &                                                         &                                                              \\
		& Pool Size=2                                                                                                                                         &                                                               &                                                          &                                                         &                                                              \\
		& Dense Layer=100                                                                                                                                     &                                                               &                                                          &                                                         &                                                              \\
		& Dense Layer=1                                                                                                                                       &                                                               &                                                          &                                                         &                                                              \\ \hline
	\end{tabular}
\caption{Comparison of method parameters for ablation experiments}
\end{table}
	
	In order to explore the relationship between MFF's ability to improve the prediction effect and MLP, a combination of MFF and LSTM and CNN are used for ablation experiments. LSTM and CNN are common deep learning neural network models, which have applications in time series prediction \cite{zeng2016self, cao2019financial}. By replacing the MLP inside MFF with CNN, LSTM, the model parameters of the comparison method in the ablation experiment are shown in the Tab.7. 
	
	The MFF module is essentially a feature extraction, and the effect is similar to the convolutional layer, but MFF is a directional feature extraction. CNN can extract effective features through the convolutional layer and the pooling layer, but this feature is not a definite feature, it will change with the change of data and training parameters, and it is not robust in prediction. LSTM solves the long-term dependence on information, but when the data passes through the MFF, the $index$ feature eliminates the context of the time series, and the learning effect will not change due to the learning order. The combination of CNN and MFF will lose information, while LSTM will lose its long-term dependence on information. Single MLP is a non-linear fitting of the original time series data, which has no advantages compared with CNN and LSTM \cite{borghi2021covid}. Under this premise, combining the MFF module with MLP would be a relatively good choice.
	
	Tab.8 and Fig.14 are the comparison of the ablation experiment. According to the experimental results, MFF performs better than single deep learning prediction method MLP, CNN and LSTM and MFF combined with CNN and LSTM.

	\begin{figure}[htbp]
		\centerline{\includegraphics[scale=0.8]{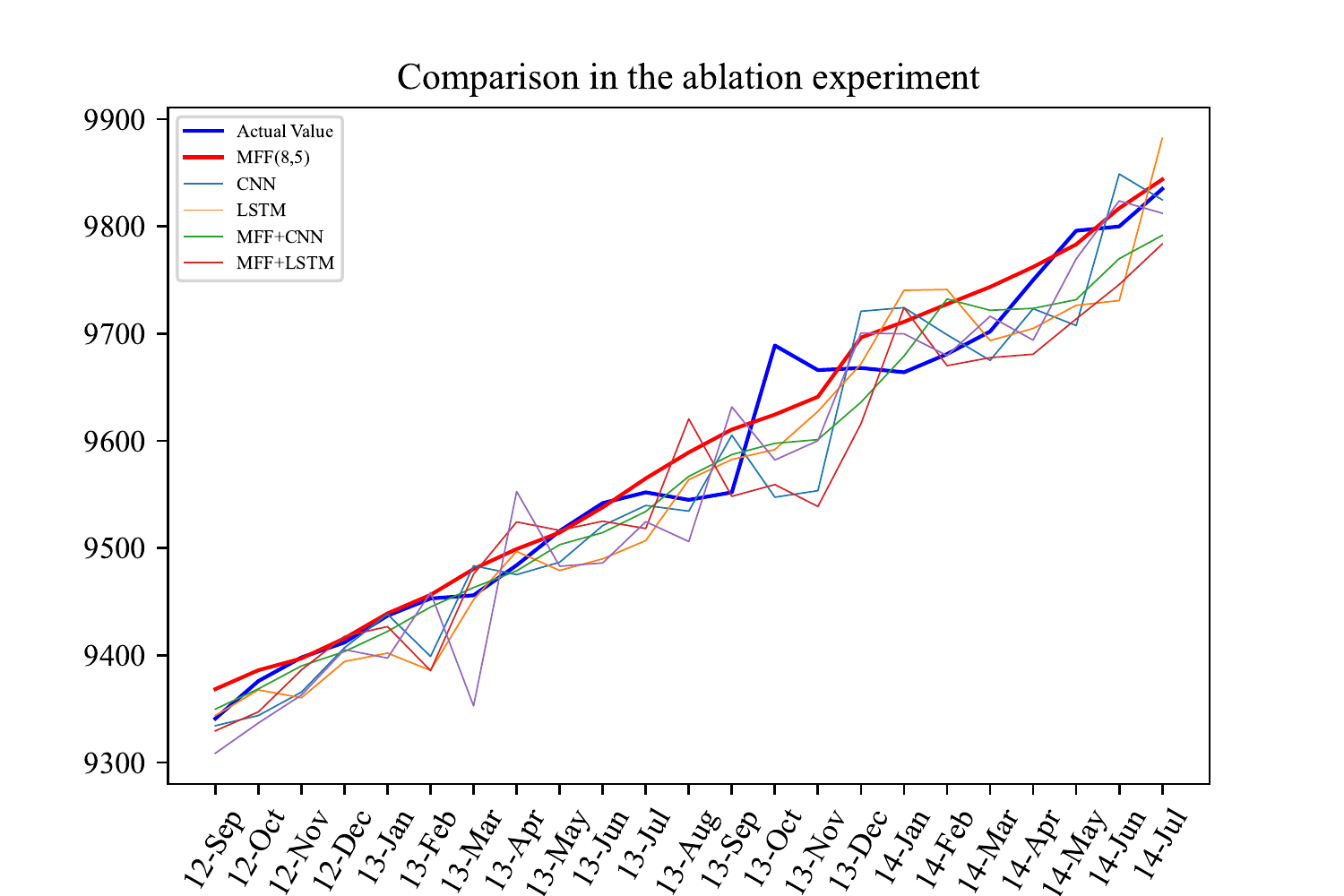}}
		\caption{Comparison in the ablation experiment}
	\end{figure}
	
	\begin{table}[htbp]
					\centering
		\begin{tabular}{cccccc}
			\hline
			& MAD              & MAPE            & SMAPE           & RMSE             & NRMSE             \\ \hline
			MFF+CNN           & 42.9095          & 0.4474          & 0.4463          & 55.9896          & 262.7687          \\
			MFF+LSTM          & 41.3319          & 0.4332          & 0.4325          & 49.7223          & 219.0653          \\
			CNN  \cite{zeng2016self}             & 38.4514          & 0.4005          & 0.4000          & 46.3639          & 199.8156          \\
			MLP \cite{borghi2021covid}              & 38.2716          & 0.3991          & 0.3982          & 51.7000          & 227.9186          \\
			LSTM  \cite{cao2019financial}            & 27.0059          & 0.2805          & 0.2801          & 34.8895          & 165.9856          \\
			\textbf{MFF(8,5)} & \textbf{22.2877} & \textbf{0.2318} & \textbf{0.2316} & \textbf{29.2458} & \textbf{131.5833} \\ \hline
		\end{tabular}
	\caption{Forecasting error in the ablation experiment}
	\end{table}

	\subsection{Additional experiment}
	In order to show the prediction effect of MFF$(M,N)$, the experimental effect of different MLP parameters $(M, N)$ will be tested here. Both M and N were tested from 1 to 20, and a total of 400 models were tested. At the same time, the top 10 models and errors of the prediction effect are shown in Tab.9.
	
	\begin{table}[htbp]
					\centering
		\begin{tabular}{ccccccc}
			\hline
			M & N  & MAD     & MAPE   & SMAPE  & RMSE    & NRMSE    \\ \hline
			3 & 8  & 19.6209 & 0.2041 & 0.2041 & 26.9679 & 121.3342 \\
			2 & 9  & 20.0718 & 0.2090 & 0.2089 & 27.1621 & 122.2081 \\
			1 & 6  & 21.1527 & 0.2204 & 0.2202 & 27.9791 & 125.8839 \\
			5 & 13 & 22.0341 & 0.2293 & 0.2292 & 29.2146 & 131.4428 \\
			3 & 16 & 22.1317 & 0.2303 & 0.2301 & 29.1754 & 131.2663 \\
			1 & 7  & 22.1498 & 0.2306 & 0.2305 & 29.2729 & 131.7052 \\
			8 & 5  & 22.2877 & 0.2318 & 0.2316 & 29.2458 & 131.5833 \\
			8 & 9  & 23.1081 & 0.2405 & 0.2402 & 29.9967 & 134.9616 \\
			1 & 20 & 23.6854 & 0.2463 & 0.2460 & 30.7693 & 138.4374 \\
			9 & 2  & 23.7177 & 0.2468 & 0.2465 & 30.5634 & 137.5111 \\ \hline
		\end{tabular}
	\caption{Top 10 models with the best prediction results of MFF(M,N)}
	\end{table}
	
	Experimental results show that the prediction performance of MFF can be further improved by trying different MLP models in the MFF. MFF has the potential for further improvement.

	\subsection{Experiment conclusion}
	In MFF, the function sequence contains the generation methods for the characteristics of multiple directions of the time sequence. Through the method of information fusion, MFF fuses different features into prediction targets. Feature fusion uses machine learning which more flexibly fuse target data based on existing data. The Adam and CLR algorithms are used for back propagation and parameter optimization of MLP, which improves the training effect while increasing the robustness and efficiency of MLP.
	
	The prediction effect of MFF is better than common statistics, machine learning and hybrid methods. Compared with deep learning CNN and LSTM neural network predictions, the directional feature learning of the MFF module determines that MLP is a relatively suitable model. After ablation experiments, it is proved that the combination of MFF module and MLP has a better prediction effect. In additional experiments, by adjusting the parameters of MFF, the prediction accuracy of MFF is further improved, and MFF has the potential to continue to improve the prediction effect. Among the five statistical error indicators in the experiment, MFF has the smallest error, indicating that MFF has high prediction accuracy.
	
	In terms of time complexity, the time complexity of the MFF module depends on the function sequence selected in MFF. In this experiment, the time complexity of the MFF module is $O(n^2)$ and the time complexity of the convolutional layer in CNN is the same. In terms of actual test time, MFF has good predictive performance.
	
	In a conclusion, MFF has a more accurate prediction effect and efficient prediction performance. At the same time, MFF has the potential to further improve forecast accuracy.

	\section{Conclusion}
 	The paper proposed the MFF method to predict CCI. By combining information fusion and machine learning, the prediction effect of MFF has been improved compared with commonly used prediction methods. The proposal of MFF has contributed to CCI and time series forecasting. In the future, MFF will continue to improve and explore time series forecasting methods based on information fusion and machine learning.

	\section*{Acknowledgment}
	This research is supported by the National Natural Science Foundation of China (No. 62003280) and Chongqing Talents: Exceptional Young Talents Project (CQYC202105031).

	\section*{Conflict of Interests}
	The authors declare that there are no conflict of interests.

	\bibliographystyle{elsarticle-num}
	\bibliography{References}

\end{document}